\documentclass[10pt,twocolumn,letterpaper]{article}

\usepackage{iccv}
\usepackage{times}
\usepackage{epsfig}
\usepackage{graphicx}
\usepackage{amsmath}
\usepackage{amssymb}
\usepackage{rotating}

\usepackage[breaklinks=true,bookmarks=false]{hyperref}

\iccvfinalcopy 

\def\iccvPaperID{****} 

\begin{document}

\title{Action Recognition by Hierarchical Mid-level Action Elements}


\author{Tian Lan$^{*}$, Yuke Zhu\thanks{indicates equal contribution} , Amir Roshan Zamir and Silvio Savarese \\
Stanford University
}

\maketitle

\begin{abstract}
Realistic videos of human actions exhibit rich spatiotemporal structures at multiple levels of granularity: an action can always be decomposed into multiple finer-grained elements in both space and time. To capture this intuition, we propose to represent videos by a hierarchy of mid-level action elements (MAEs), where each MAE corresponds to an action-related spatiotemporal segment in the video. We introduce an unsupervised method to generate this representation from videos. Our method is capable of distinguishing action-related segments from background segments and representing actions at multiple spatiotemporal resolutions.  

Given a set of spatiotemporal segments generated from the training data, we introduce a discriminative clustering algorithm that automatically discovers MAEs at multiple levels of granularity. We develop structured models that capture a rich set of spatial, temporal and hierarchical relations among the segments, where the action label and multiple levels of MAE labels are jointly inferred. The proposed model achieves state-of-the-art performance in multiple action recognition benchmarks. Moreover, we demonstrate the effectiveness of our model in real-world applications such as action recognition in large-scale untrimmed videos and action parsing.



\end{abstract}

\section{Introduction}
\begin{figure}
\centering
\includegraphics[width=\columnwidth]{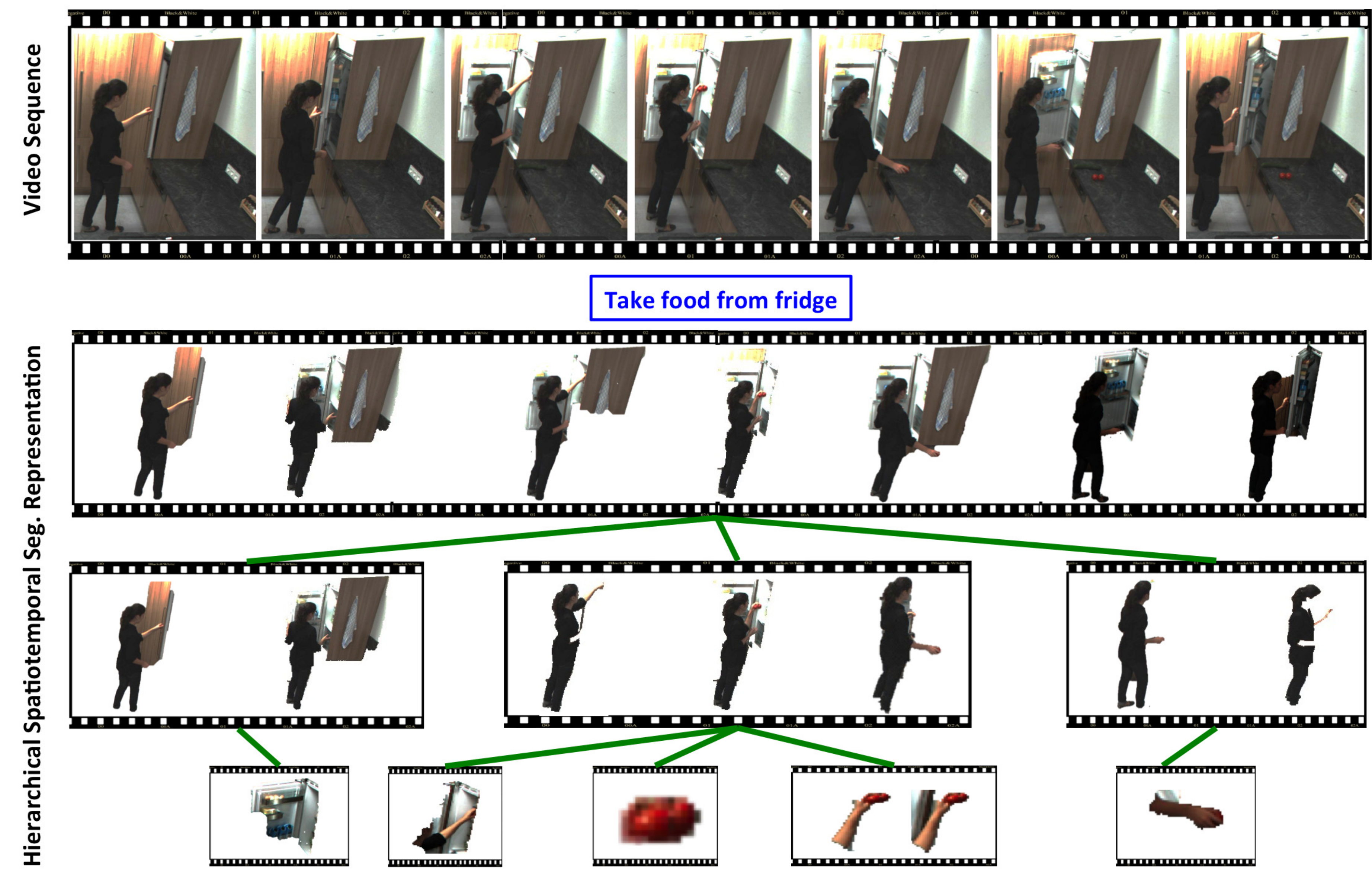} 
\caption{{\bf A representation of {\em hierarchical spatiotemporal segments} for action.} Our method automatically discovers representative and discriminative mid-level action elements for a given action class. These elements are encoded in the spatiotemporal segments which usually cover different aspects of an action at different levels of granularity, ranging from an entire action sequence, which comprises the actor along with the objects the actor interacts with (the first row of the hierarchy), to the action elements such as fine-grained body part movements and objects (the last row).
\vspace{-4mm}}
\label{fig:open}
\end{figure}

In this paper we address the problem of learning models of human actions and using these models for recognizing and parsing human actions from videos. This is a very challenging problem. Most of the human actions are complex spatial-temporal hierarchical processes. Consider, for instance, the action in Fig.~\ref{fig:open}. This is composed of a collection of spatiotemporal processes ranging from the entire action sequence, ``taking food from fridge'' to simple elementary actions such as ``stretching arm'' or ``grasping a tomato''. Each of these actions is often characterized by a complex distribution of motion segments~(e.g.~\emph{open} and \emph{close}), objects~(e.g.~\emph{fridge} and \emph{food}), body parts~(e.g.~\emph{arm}) along with their interactions~(e.g.~\emph{grasp a tomato}). Thus, in order to achieve a full understanding of the action that takes place in a scene, one must recognize and parse this complex structure of \emph{mid-level action elements} (MAEs) at different levels of semantic and spatial-temporal resolution.

Most of the existing methods cannot do this. A large body of work focuses on associating the entire video clip with a single class label from a pre-defined set of action categories (e.g., ``take food from fridge'' versus ``cook food") (Fig.~\ref{fig:open})~\cite{laptev07_iccv, laptev08_cvpr, wang13_iccv} -- essentially, a video classification problem. Methods such as~\cite{hoai11_cvpr, niebles10_eccv} do propose methodologies for temporally segmenting or parsing the action (e.g., ``take food from fridge") into a sequence of sub-action labels (e.g., open fridge, grasp food, close fridge) but cannot organize these sub-actions into hierarchical structures of MAEs such as the one in Fig.~\ref{fig:open}. Critically, most of these methods assume that the fine-grained action labels or their temporal structures are pre-defined or hand-specified by an expert as opposed to be automatically inferred from the videos in a data-driven fashion. This assumption prevents such methods from scaling up to a large number of complex actions. Finally, a portion of previous research focuses on modeling an action by just capturing the spatial-temporal characteristics of the actor~\cite{efros03_iccv, shechtman05_cvpr, lan11_iccv, raptis13_cvpr} whereby neither the objects nor the background the actor interacts with are used to better contextualize the classification process.  Other methods~\cite{klaser08_bmvc, wang13_iccv, kovashka10_cvpr, ryoo09_iccv} do propose a holistic representation for activities which inherently captures some degree of background context in the video, but are unable to spatially localize or segment the actors or relevant objects.

In this work, we propose a model that is capable of modeling complex actions as a collection of {\em mid-level action elements} (MAEs) that are organized in a hierarchical way. Compared to previous approaches, our framework enables: 1) {\bf Multi-resolution reasoning} -- videos can be decomposed into a hierarchical structure of spatiotemporal MAEs at multiple scales; 2) {\bf Parsing capabilities} -- actions can be described (parsed) as a rich collection of spatiotemporal MAEs that capture different characteristics of the action ranging from small body motions, objects to large pieces of volumes containing person-object interactions. These MAEs can be spatially and temporally localized in the video sequence; 3) {\bf Data-driven learning} -- the hierarchical structure of MAEs as well as the their labels do not have to be manually specified, but are learnt and discovered automatically using a newly proposed weakly-supervised agglomerative clustering procedure. Note that some of the MAEs might have clear semantic meanings (see Fig.~\ref{fig:open}), while others might correspond to random but discriminative spatiotemporal segments. In fact, these MAEs are learnt so as to establish correspondences between videos from the same action class while maximizing their discriminative power for different action classes. Our model has achieved state-of-the-art results on multiple action recognition benchmarks and is capable of recognizing actions from large-scale untrimmed video sequences.

\section{Related Work}
\label{sec:rw}
The literature on human action recognition is immense. we refer the readers to the recent survey~\cite{aggarwal11_acm}. In the following, we only review the related work closely to our work.

{\bf Space-time segment representation:} Representing actions as 2D+t tubes is a common strategy for action recognition~\cite{blank05_iccv, brendel11_iccv, ma13_iccv}. Recently, there are works that use hierarchical spatiotemporal segments to capture the multi-scale characteristics of actions~\cite{brendel11_iccv, ma13_iccv}. Our representation differs in that we can discriminatively discover the \emph{mid-level action elements} (MAEs) from a pool of region proposals. 

{\bf Temporal action localization:} While most action recognition approaches focus on classifying trimmed video clips~\cite{laptev08_cvpr, efros03_iccv, ryoo09_iccv}, there are works that attempt to localize action instances from long video sequences~\cite{duchenne09_iccv, hoai11_cvpr, laptev07_iccv, bojanowski14_eccv, pirsiavash14_cvpr}. In~\cite{pirsiavash14_cvpr}, a grammar model is developed for localizing action and (latent) sub-action instances in the video. Our work considers a more detailed parsing at both space and time, and at different semantic resolutions.

{\bf Hierarchical structure:} Hierarchical structured models are popular in action recognition due to its capability in capturing the multi-level granularity of human actions~\cite{tang12_cvpr, lan14_eccv, lan14_vs, raptis13_cvpr}. We follow a similar spirit by representing an action as a hierarchy of MAEs. However, most previous works focus on classifying single-action video clips where they treat these MAEs as latent variables. Our method localizes MAEs at both spatial and temporal extent.

{\bf Data-driven action primitives:} Action primitives are discriminative parts that capture the appearance and motion variations of the action~\cite{niebles10_eccv, yao10b_cvpr, jain13_cvpr, lan14_vs}. Previous representations of action primitives such as interest points~\cite{yao10b_cvpr}, spatiotemporal patches~\cite{jain13_cvpr} and video snippets~\cite{niebles10_eccv} typically lack multiple levels of granularity and structures. In this work, we represent action primitives as MAEs, which are capable of capturing different aspects of actions ranging from the fine-grained body part segments to the large chunks of human-object interactions. A rich set of spatial, temporal and hierarchical relations between the MAEs are also encoded. Both the MAE labels and the structures of MAEs are discovered in a data-driven manner.

Before diving into details, we first give an overview of our method. 1) {\em Hierarchical spatiotemporal segmentation}. Given a video, we first develop an algorithm to automatically parse the video into a hierarchy of spatiotemporal segments~(see Fig.~\ref{fig:pipeline}). We run this algorithm for each video independently, and in this way, each video is represented as a spatiotemporal segmentation tree (Section~\ref{sec:hierarchy}). 2) {\em Learning.} Given a set of spatiotemporal segmentation trees (one tree per video) in training, we propose a graphical model that  captures the hierarchical dependencies of MAE labels at different levels of granularity. We consider a weakly supervised setting, where only the action label is provided for each training video, while the MAE labels are discriminatively discovered by clustering the spatiotemporal segments. The structure of the model is defined by the spatiotemporal segmentation tree where inference can be carried out efficiently~(Section~\ref{sec:model}). 3) {\em Recognition and parsing.} A new video is represented by the spatiotemporal segmentation tree. We run our learned models on the tree for recognizing the actions and parsing the videos into MAE labels at different spatial, temporal and semantic resolutions.




\section{Action Proposals: Hierarchical Spatiotemporal Segments}
\label{sec:hierarchy}
\begin{figure}
\centering
\includegraphics[width=\columnwidth]{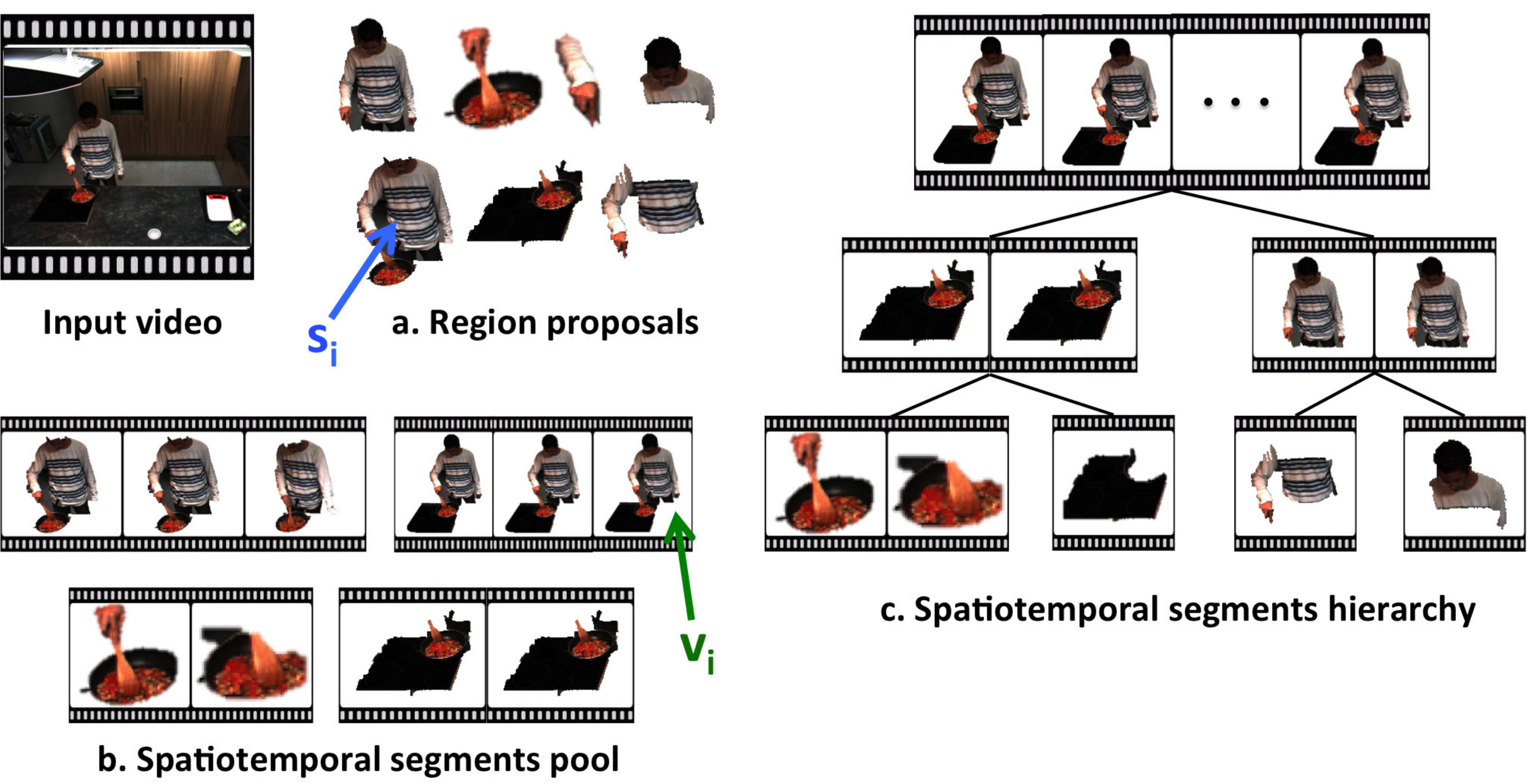} 
\caption{{\bf Constructing the spatiotemporal segment hierarchy.} (a) Given a video, we first generate action-related region proposals for each frame. (b) Then, we cluster these proposals to produce a pool of spatiotemporal segments. (c) The last step is to agglomeratively cluster the spatiotemporal segments into a hierarchy.}
\label{fig:pipeline}
\vspace{-4mm}
\end{figure}

In this section, we describe our method for generating a hierarchy of action-related spatiotemporal segments from a video. Our method is unsupervised, i.e. during training, the spatial locations of the persons and objects are not annotated. Thus, it is important that the our method can automatically extract the action-related spatiotemporal segments such as actors, body parts and objects from the video. An overview of the method is shown in Fig.~\ref{fig:pipeline}. 

Our method for generating action proposals includes three major steps. {\bf A. Generating action-related spatial segments.} We initially generate a diverse set of region proposals using the method of~\cite{endres10_eccv}. This method works on a single frame of video, and returns a large number of segmentation masks that are likely to contain objects or object parts. We then score each region proposal using both appearance and motion cues, and we look for regions that have generic object-like appearance and distinct motion patterns relative to their surroundings. We further prune the background region proposals by training an SVM using the top scored region proposals as positive examples along with patches randomly sampled from the background as negative examples. The region proposals with scores above a threshold~($-1$) are considered as action-related spatial segments. {\bf B. Obtaining the spatiotemporal segment pool.} Given the action-related spatial segments for each frame, we seek to compute ``tracklets'' of these segments over time to construct the spatiotemporal segments. We perform spectral clustering based on the color, shape and space-time distance between pairs of spatial segments to produce a pool of spatiotemporal segments. In order to maintain the purity of each spatiotemporal segment, we set the number of clusters to a reasonably large number. The pool of spatiotemporal segments correspond to the action proposals at the finest scale (bottom of Fig.~\ref{fig:open}). {\bf C. Constructing the hierarchy.} Starting from the initial set of fine-grained spatiotemporal segments, we agglomeratively cluster the most similar spatiotemporal segments into super-spatiotemporal segments until a single super-spatiotemporal segment is left. In this way, we produce a hierarchy of spatiotemporal segments for each video that forms a tree structure, i.e. action proposals at different levels of granularity. Due to space constraints, we refer the details of the method to the supplementary material.


\section{Hierarchical Models for Action Recognition and Parsing}
\label{sec:model}
So far we have explained how to parse a video into a tree of spatiotemporal segments. We run this algorithm for each video independently, and in this way, each video is represented as a tree of spatiotemporal segments. Our goal is to assign each of these segments to a label so as to form a \emph{mid-level action element} (MAE). We consider a weakly supervised setting. During training, only the action label is provided for each video. We discover the MAE labels in an unsupervised way by introducing a discriminative clustering algorithm that assigns each spatiotemporal segment to an MAE label~(Section~\ref{sec:primitive_discover}). In Section~\ref{sec:model_formulation}, we introduce our models for action recognition and parsing, which are able to capture the hierarchical dependencies of the MAE labels at different levels of granularity. 

\begin{figure}
\centering
\includegraphics[width=0.9\columnwidth]{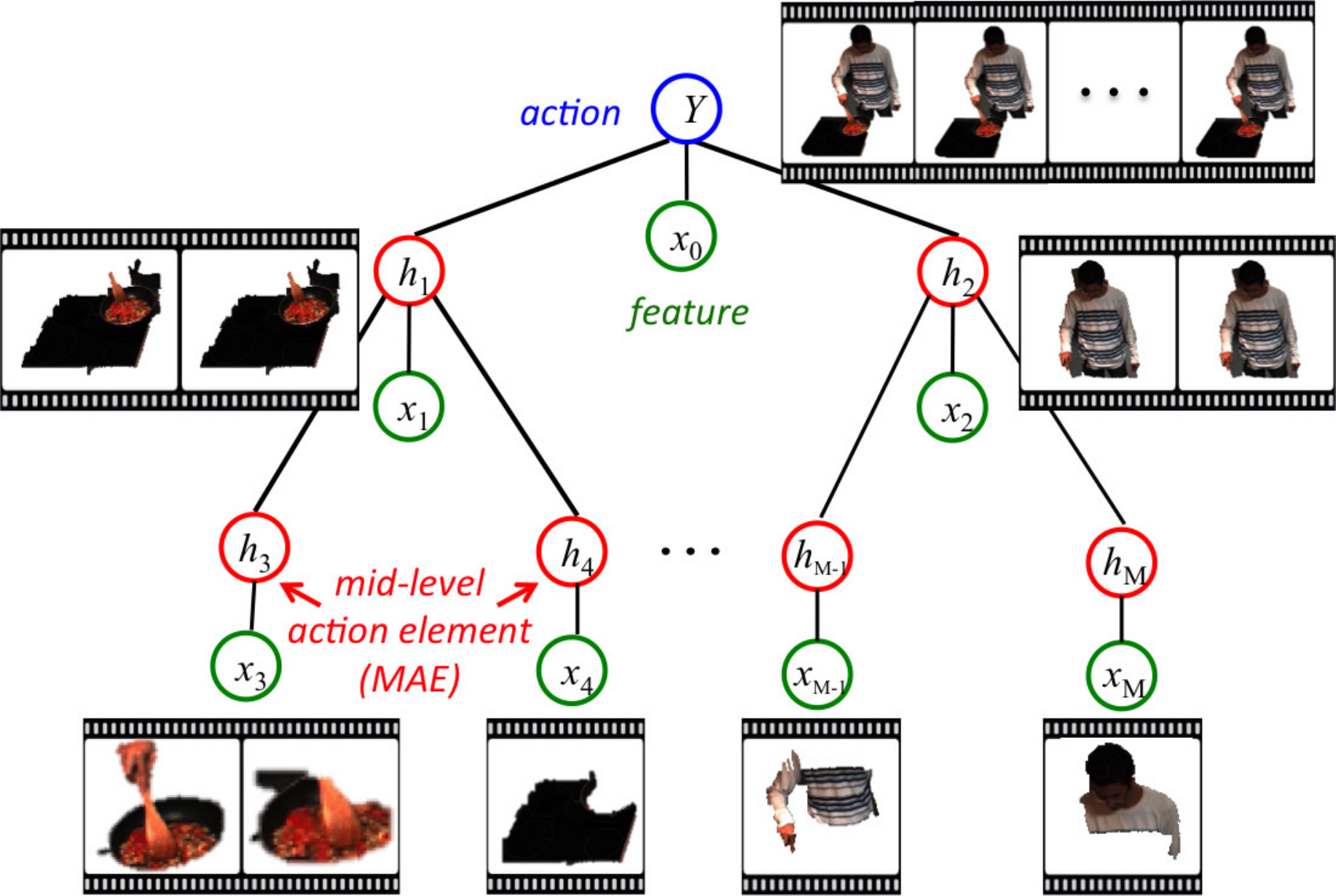} 
\caption{{\bf Graphical illustration of the model.} In this example, we adopt the spatiotemporal hierarchy in Fig.~\ref{fig:pipeline}~(c). The MAE labels are the red circles. The green circles are the features of each spatiotemporal segment, and the the blue circle is the action label.}
\label{fig:graph}
\vspace{-5mm}
\end{figure}

We start by describing the notations. Given a video $V_n$, we first parse it into a hierarchy of spatiotemporal segments, denoted by $V_n = \{ v_i: i = 1,\ldots,M_n\}$ following the procedure introduced in Section~\ref{sec:hierarchy}. We extract features $X_n$ from these spatiotemporal segments in the form of $X_n = \{ x_i: i = 0, 1,\ldots, M_n\}$, where $x_0$ is the root feature vector, computed by aggregating the feature descriptors of all spatiotemporal segments in the video, and $x_i~(i=1,\dots,M_n)$ is the feature vector extracted from the spatiotemporal segment $v_i$~(see Fig.~\ref{fig:graph}). 

During training, each video $V_n$ is annotated with an action label $Y_n\in\mathcal{Y}$ and $\mathcal{Y}$ is the set of all possible action labels. We denote the MAE labels in the video as $H_n=\{h_i: i=1,\ldots,M_n\}$, where $h_i\in\mathcal{H}$ is the MAE label of the spatiotemporal segment $v_i$ and $\mathcal{H}$ is the set of all possible MAE labels~(see Fig.~\ref{fig:graph}). For each training video, the MAE labels $H_n$ are automatically assigned to clusters of spatiotemporal segments by our discriminative clustering algorithm (Section~\ref{sec:primitive_discover}). The hierarchical structure above can be compactly described using the notation $\mathcal{G}_n=(\mathcal{V}_n, \mathcal{E}_n)$,  where a vertex $v_i\in\mathcal{V}_n$ denotes a spatiotemporal segment, and an edge $(v_i,v_j)\in\mathcal{E}_n$ represents the interaction between a pair of spatiotemporal segments. In the next section, we describe how to automatically assign MAE labels to clusters of spatiotemporal segments. 


\subsection{Discovering Mid-level Action Elements (MAEs)}
\label{sec:primitive_discover}
Given a set of training videos with action labels, our goal is to discover the MAE labels $\mathcal{H}$ by assigning the clusters of \emph{spatiotemporal segments} (Section~\ref{sec:hierarchy}) to the corresponding cluster indices. Consider the example in Fig.~\ref{fig:open}, the input video is annotated with an action label ``take food from fridge'' in training, and the MAEs should describe the action at different resolutions ranging from the fine-grained action and object segments~(e.g. fridge, tomato, grab) to the higher-level human-object interactions~(e.g. open fridge, close fridge). These MAE labels are not provided in training, but are automatically discovered by a discriminative clustering algorithm on a per-category basis. That means the MAEs are discovered by clustering the spatiotemporal segments from all the training videos within each action class. The MAEs should satisfy two key requirements: 1) {\em inclusivity} - MAEs should cover all, or at least most, variations in the appearance and motion of the spatiotemporal segments in an action class; 2) {\em discriminability} - MAEs should be useful to distinguish an action class from others.

Inspired by the recent success of discriminative clustering in generating mid-level concepts~\cite{singh12_eccv}, we develop a two-step discriminative clustering algorithm to discover the MAEs. 1) \textit{Initialization:} we perform an initial clustering to partition the spatiotemporal segments into a large number of homogeneous clusters, where each cluster contains segments that are highly similar in appearance and shape. 2) \textit{Discriminative algorithm:} a discriminative classifier is trained for each cluster independently. Based on the discriminatively-learned similarity, the visually consistent clusters will then be merged into mid-level visual patterns (i.e. MAEs). The discriminative step will make sure that each MAE pattern is different enough from the rest. The two-step algorithm is explained in details below.

{\bf Initialization.} We run standard spectral clustering on the feature space of the spatiotemporal segments to obtain the initial clusters. We define a similarity between every pair of spatiotemporal segments $v_i$ and $v_j$ extracted from all of the training videos of the same class: $K(v_i, v_j) = \exp(-d_{bow}(v_i, v_j)-d_{spatial}(v_i, v_j))$, where $d_{bow}$ is the histogram intersection distance on the BoW representations of the dense trajectory features~\cite{wang13_iccv}; and $d_{spatial}$ denotes the Euclidean distance between the averaged bounding boxes of spatiotemporal segments $v_i$ and $v_j$ in terms of four cues: x-y locations, height and width. In order to keep the purity of each cluster, we set the number of clusters quite high, producing around $50$ clusters per action. We remove clusters with less than $5$ spatiotemporal segments. 

{\bf Discriminative Algorithm.} Given an initial set of clusters, we train a linear SVM classifier for each cluster on the BoW feature space. We use all spatiotemporal segments in the cluster as positive examples, and negative examples are spatiotemporal segments from other action classes. For each cluster, we run the trained discriminative classifier on all other clusters in the same action class. We consider the top $K$ scoring detections of each classifier. We define the affinity between the initial clusters $c_i$ and $c_j$ as the frequency that classifier $i$ and $j$ fire on same cluster.

\begin{figure} 
\centering
{\setlength{\tabcolsep}{0.5pt}
{\bf Take out from oven}\\
\begin{tabular}{cccccc}
\begin{sideways} \makebox[0.5in][c]{\footnotesize{MAE 1}} \end{sideways} & ~ 
\includegraphics[width=0.19\columnwidth]{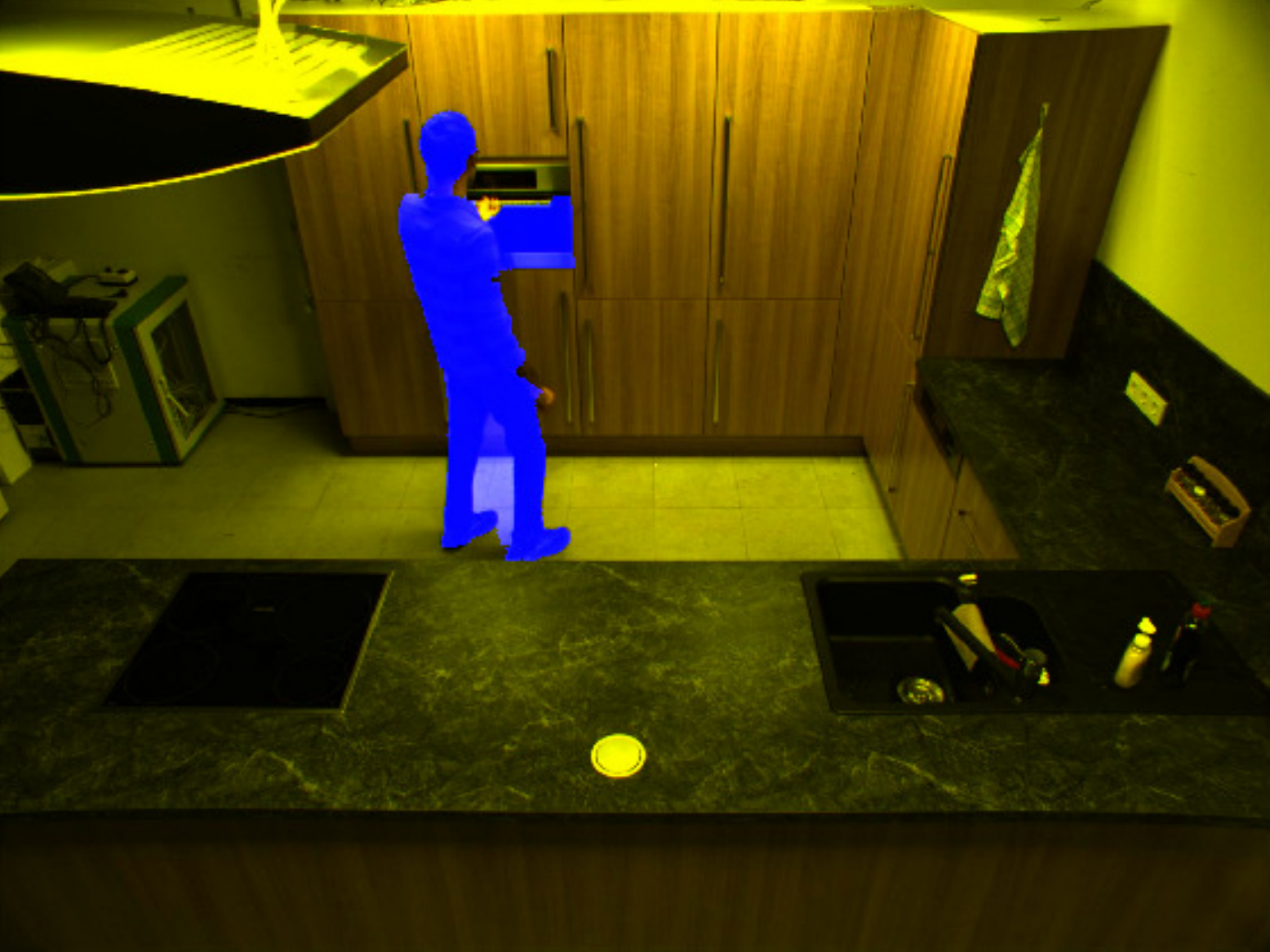} &
\includegraphics[width=0.19\columnwidth]{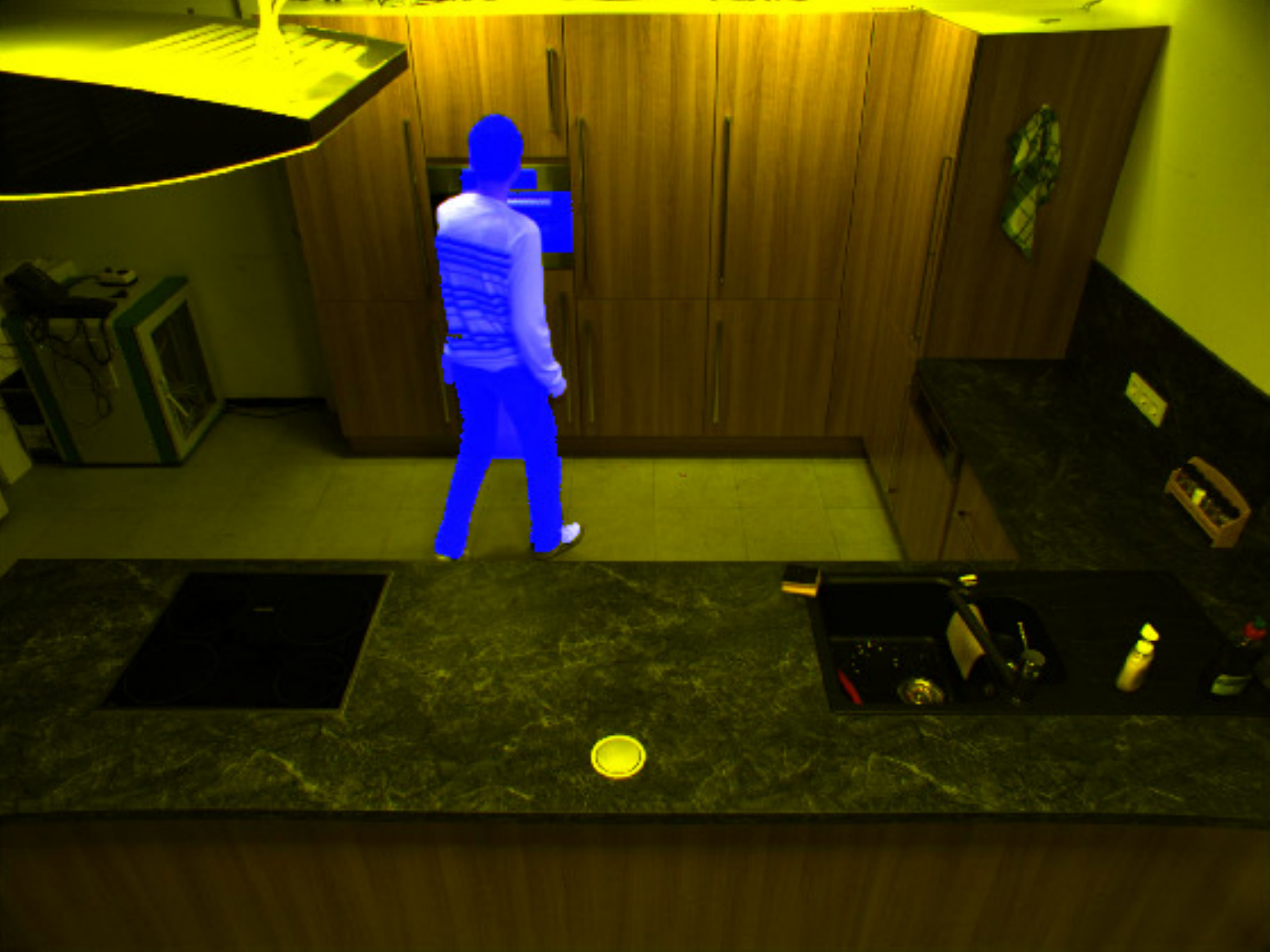} &
\includegraphics[width=0.19\columnwidth]{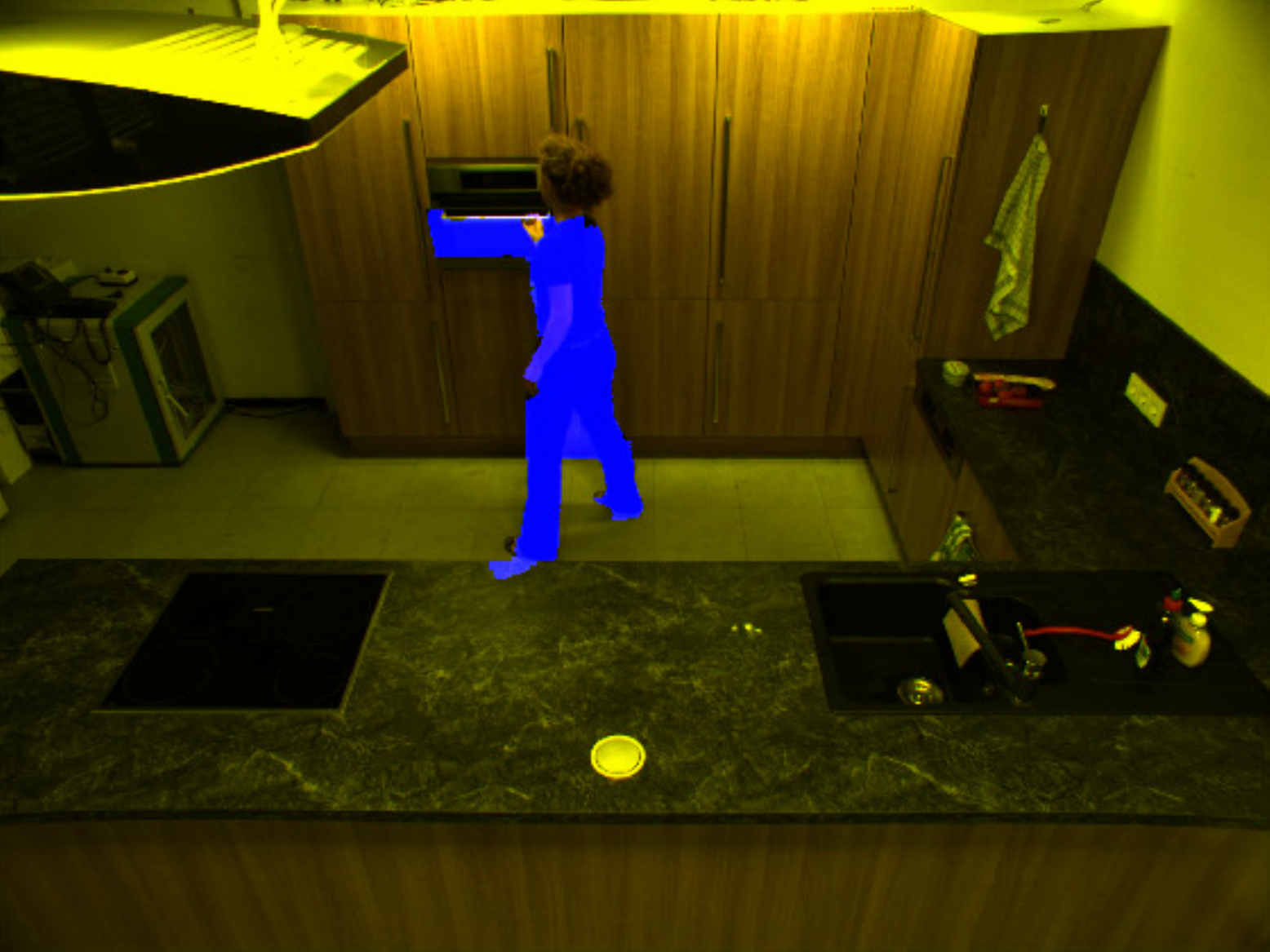} &
\includegraphics[width=0.19\columnwidth]{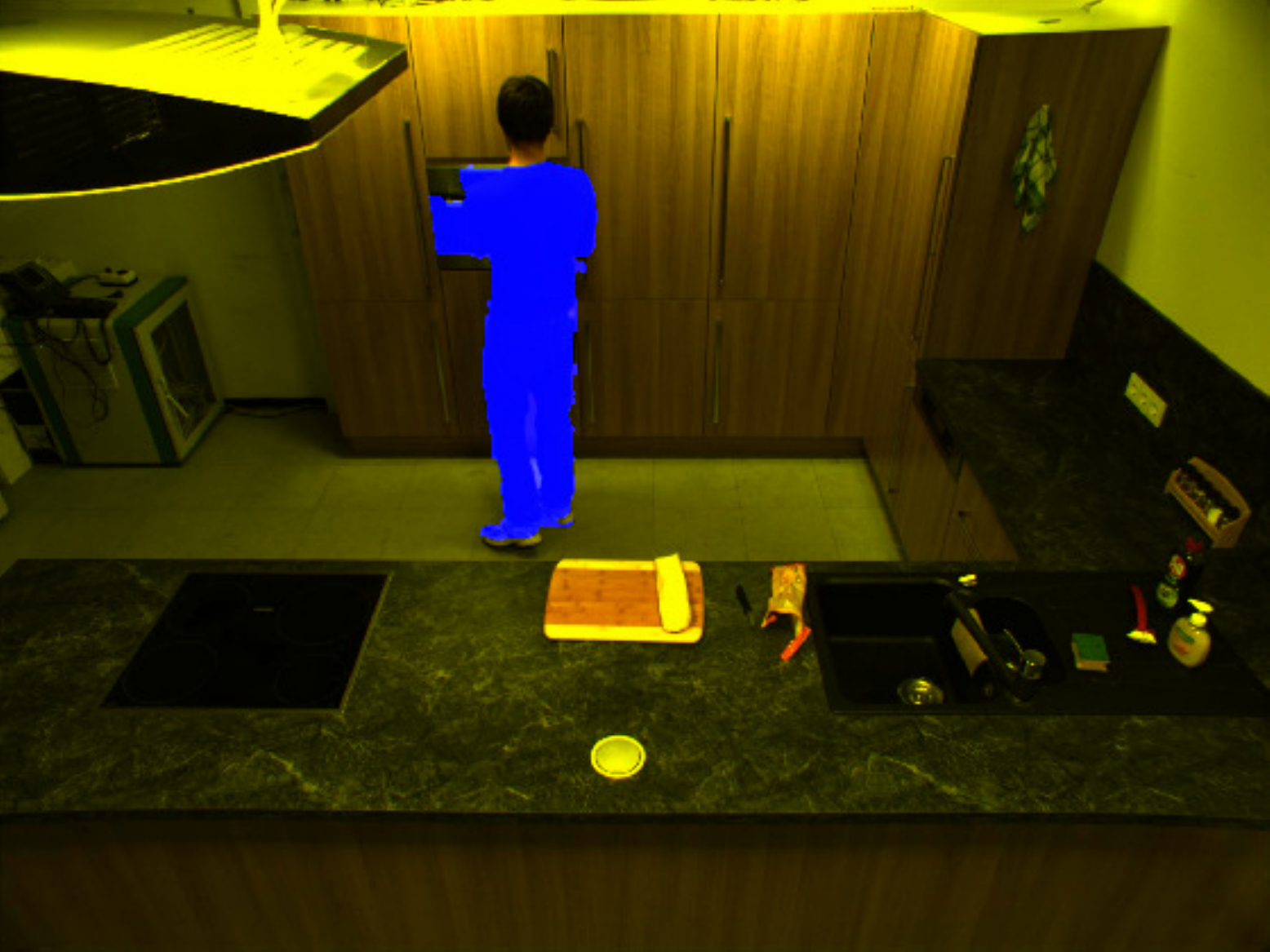} &
\includegraphics[width=0.19\columnwidth]{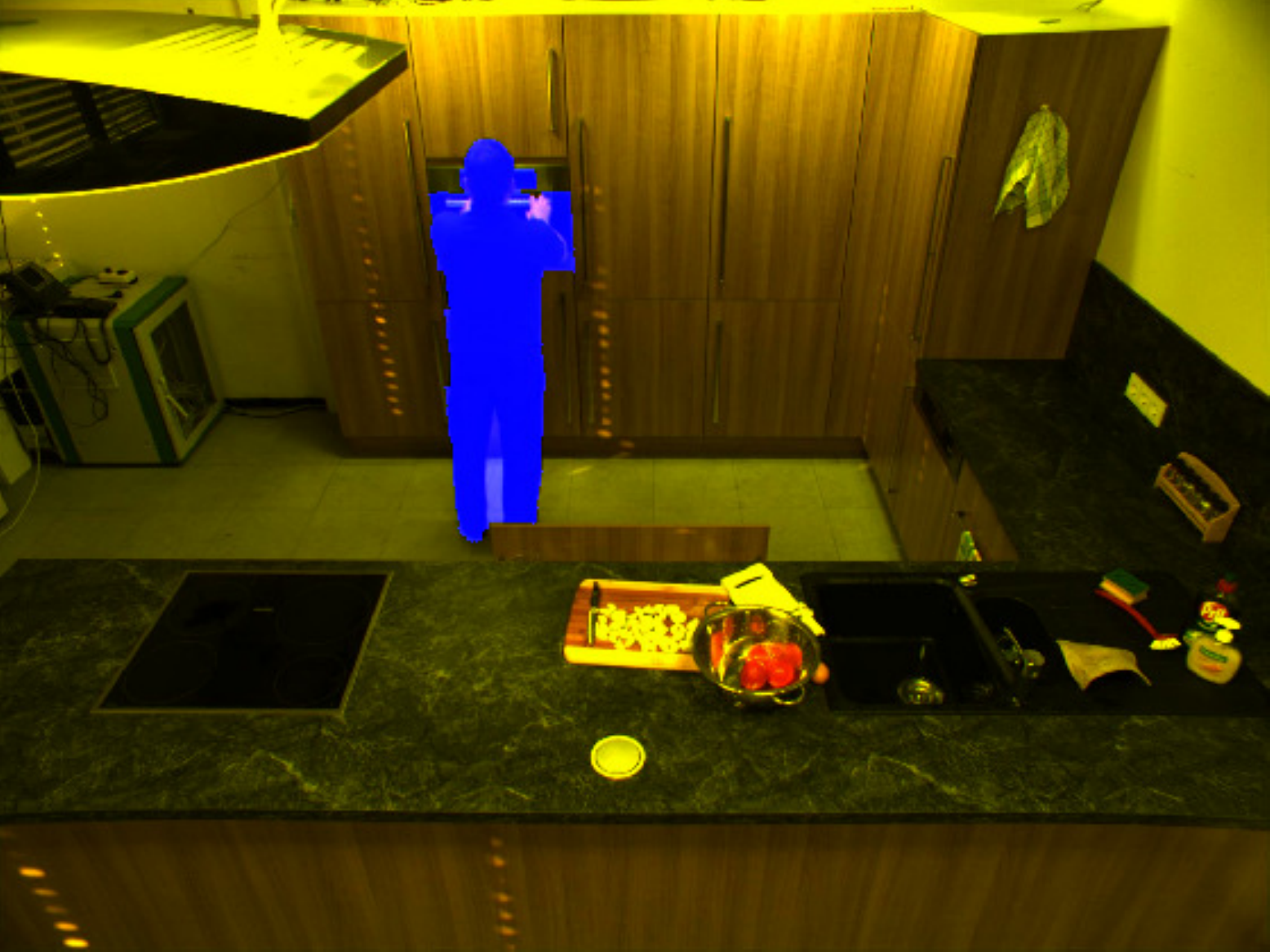} \\
\begin{sideways} \makebox[0.5in][c]{\footnotesize{MAE 2}} \end{sideways} & ~ 
\includegraphics[width=0.19\columnwidth]{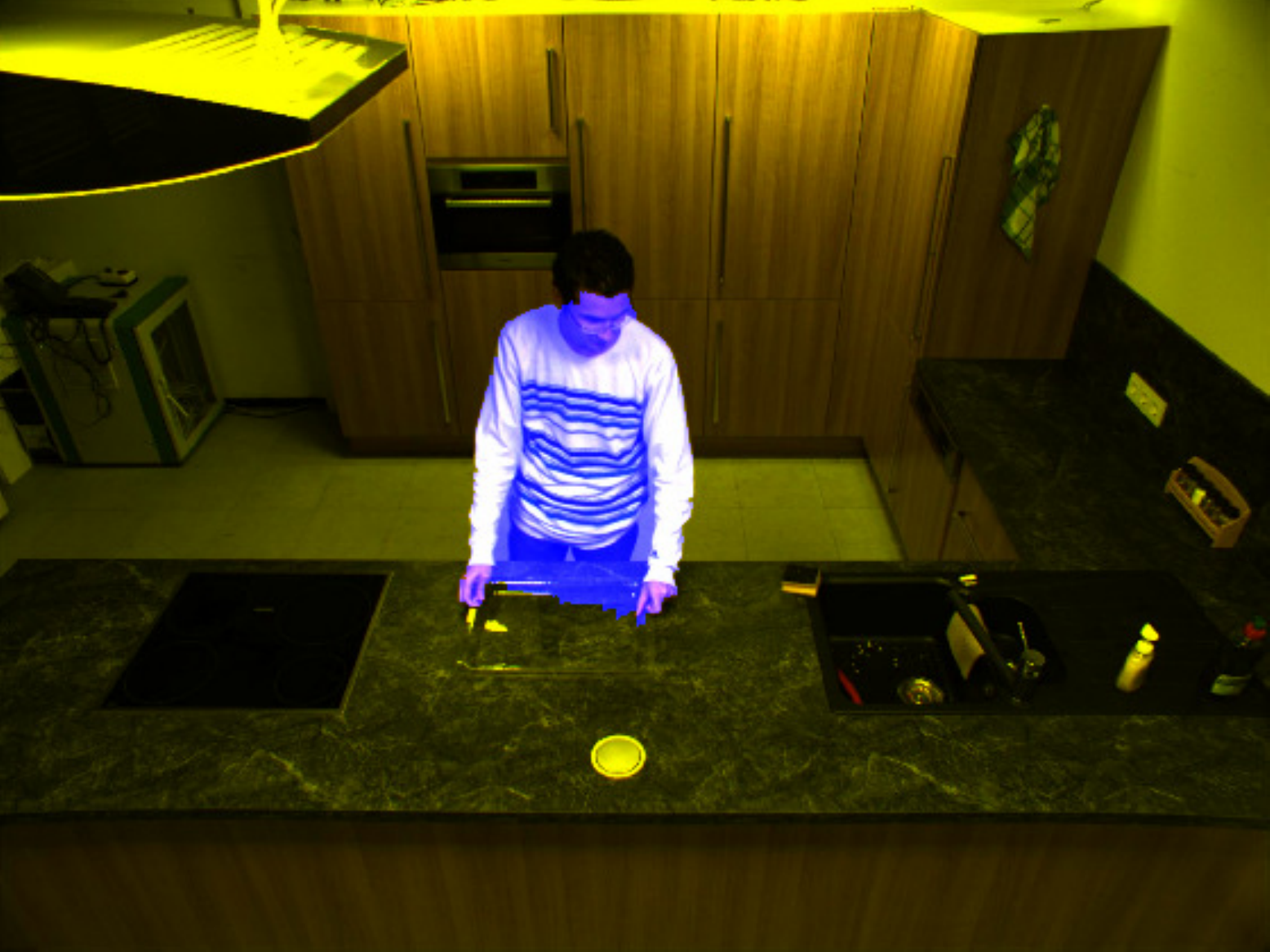} &
\includegraphics[width=0.19\columnwidth]{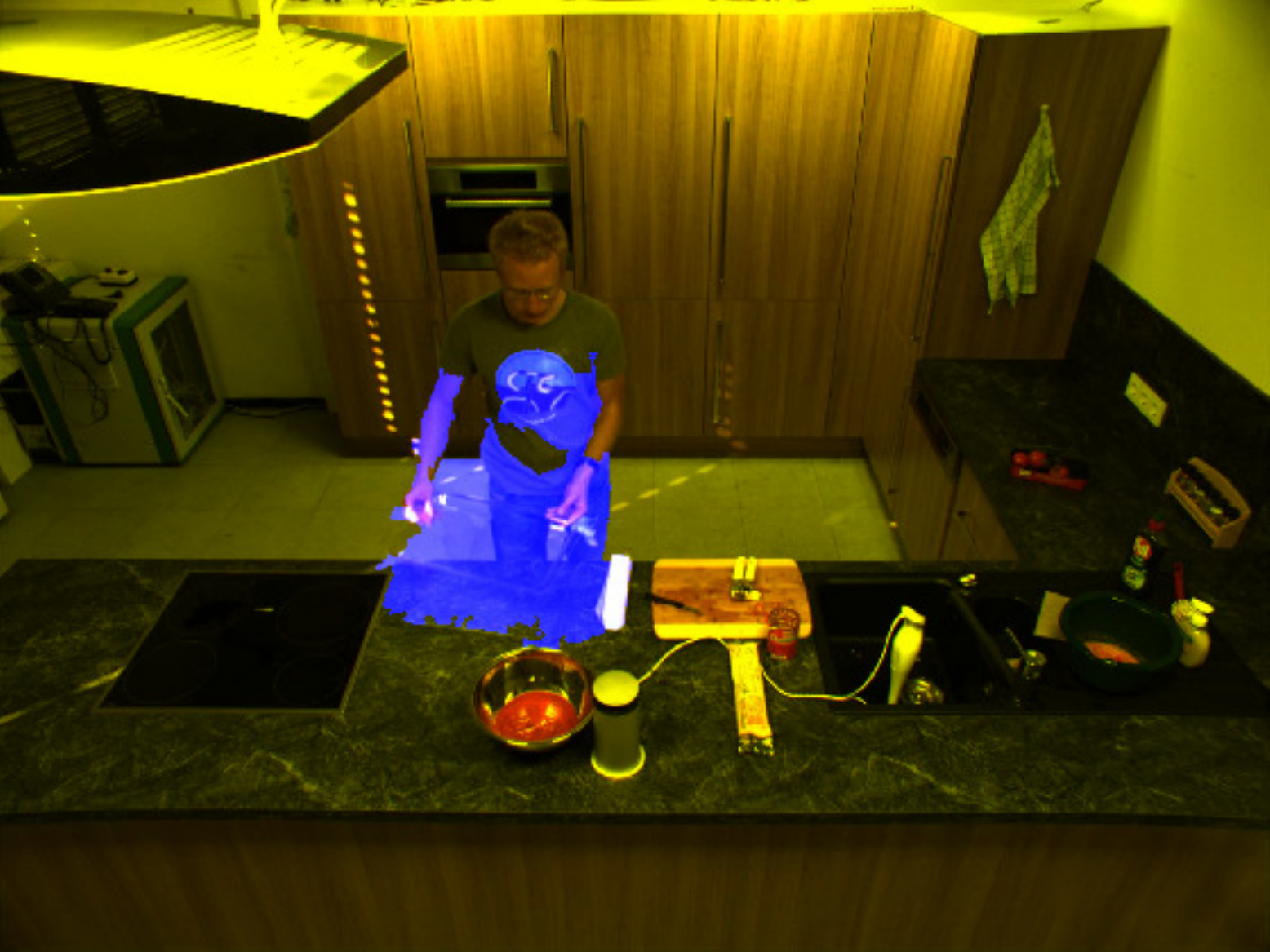} &
\includegraphics[width=0.19\columnwidth]{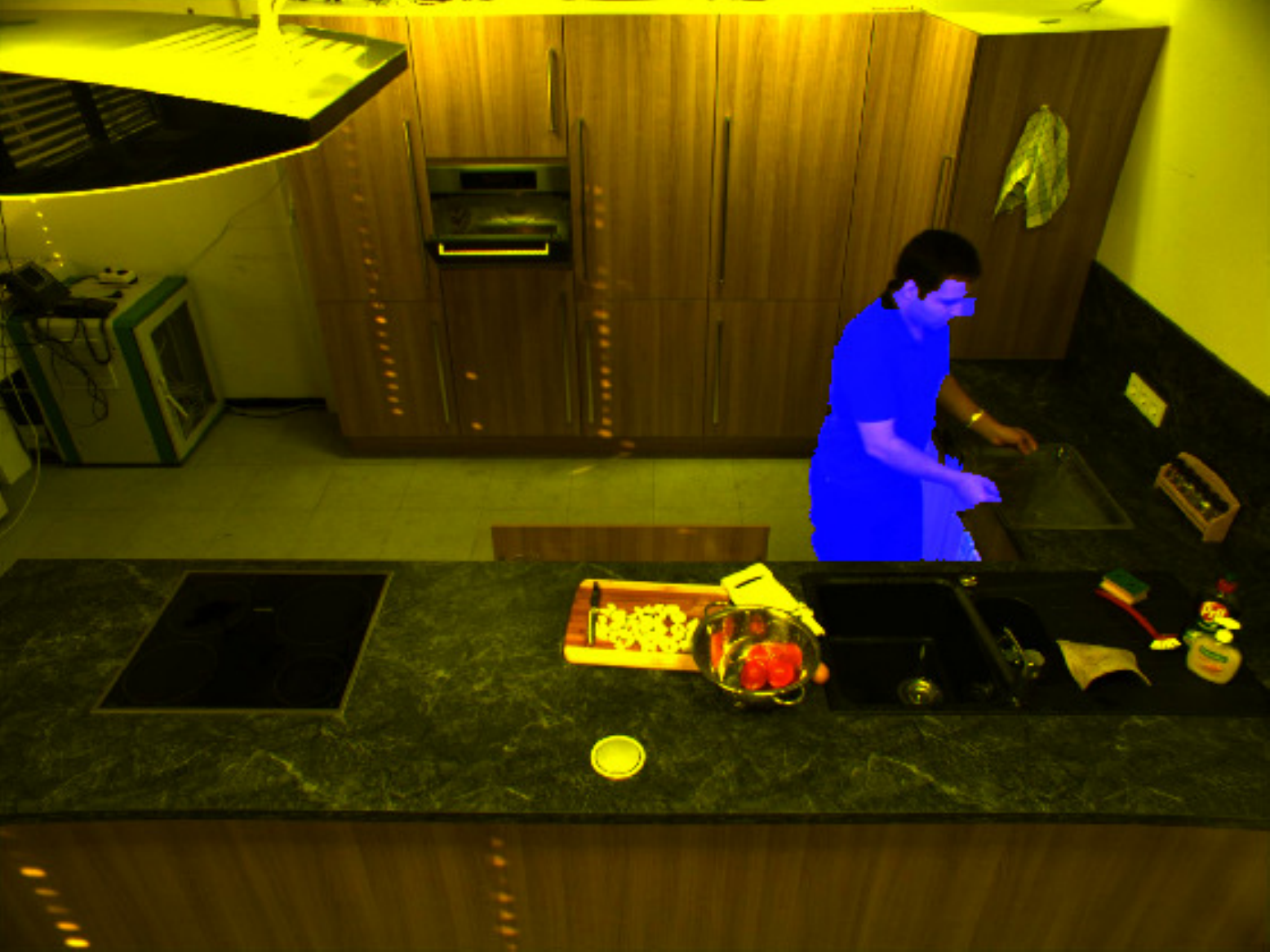} &
\includegraphics[width=0.19\columnwidth]{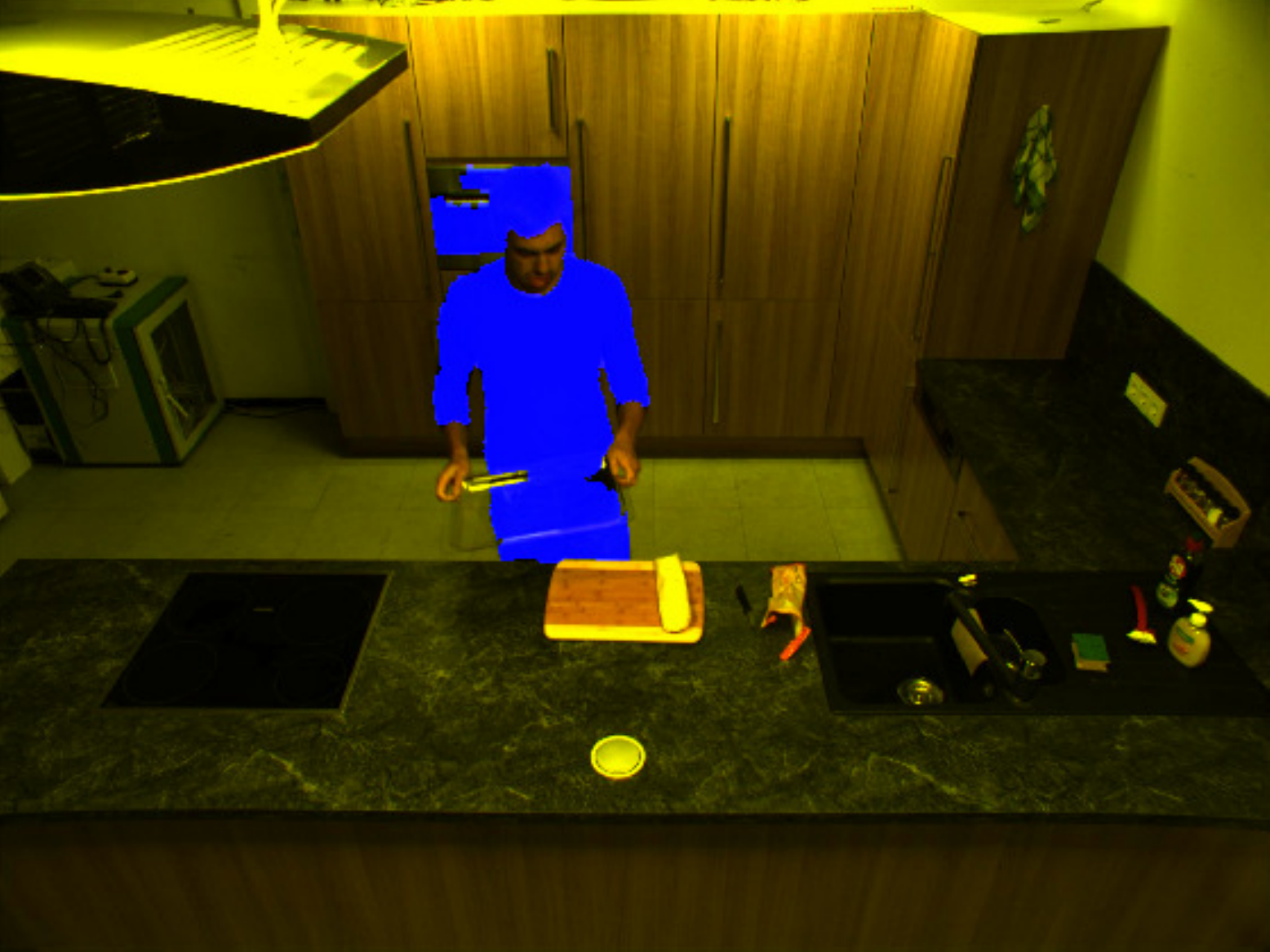} &
\includegraphics[width=0.19\columnwidth]{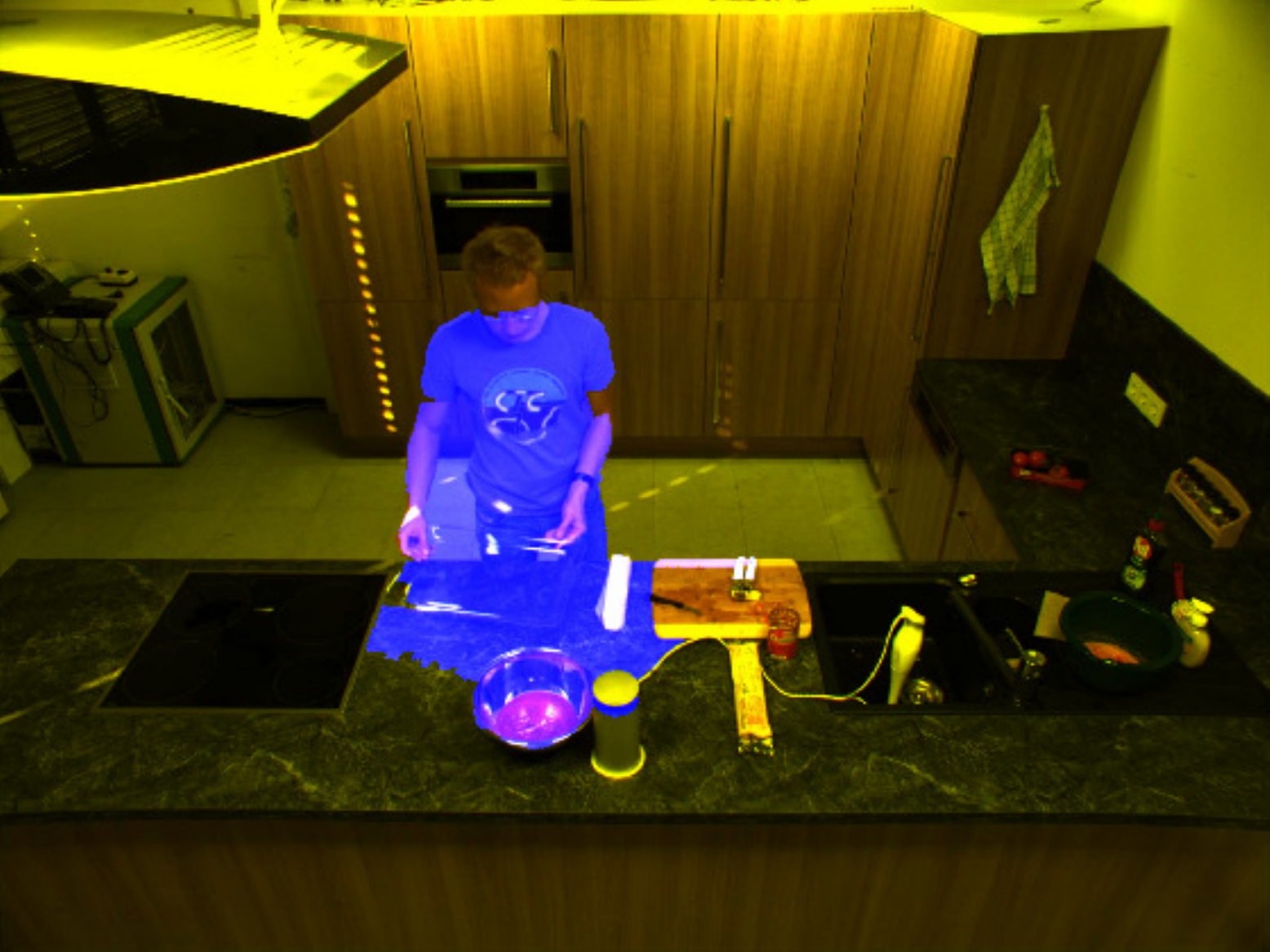} \\
\end{tabular}
}
\caption{{\bf Visualization of Mid-level Action Elements (MAEs).} The figure shows two clusters~(i.e. two MAEs) from the action category ``take out from oven''. Each image shows the first frame of a spatiotemporal segment and we only visualize five examples in each MAE. The two clusters capture two different temporal stages of ``take out from oven''. More visualizations are available in the supplementary material.} 
\label{fig:vis_action_primitive}
\end{figure}

For each action class, we compute the pairwise affinities between all initial clusters, to obtain the affinity matrix. Next we perform spectral clustering on the affinity matrix of each action independently to produce the MAE labels. In this way, the spatiotemporal segments in the training set are automatically grouped into clusters in a discriminative way, where the index of each cluster corresponds to an MAE label $h_i\in\mathcal{H}$, where $\mathcal{H}$ denotes the set of all possible MAE labels. We visualize the example MAE clusters in Fig.~\ref{fig:vis_action_primitive}. 

\subsection{Model Formulation}
\label{sec:model_formulation}
For each video, we have a different tree structure $\mathcal{G}_n$ obtained from the spatiotemporal segmentation algorithm~(Section~\ref{sec:hierarchy}). Our goal is to jointly model the compatibility between the input feature vectors $X_n$, and the action label and MAE labels ($Y_n$ and $H_n$), as well as the dependencies between pairs of MAE labels. We achieve this by using the following potential function: 
\begin{align}
\label{eq:pot_func}
&S_{V_n}(X_n, Y_n, H_n) = \sum_{i\in\mathcal{V}_n}\alpha_{h_i}^\top x_i + \sum_{i\in\mathcal{V}_n}b_{Y_n,h_i} + \sum_{\left(i,j\right)\in\mathcal{E}_n}b_{h_i,h_j}' \nonumber\\
&+\sum_{\left(i,j\right)\in\mathcal{E}}\beta_{h_i,h_j}^\top d_{ij} + \eta_{Y_n}^\top x_0 
\end{align}

{\bf MAE Model $\alpha_{h_i}^\top x_i$}: This potential captures the compatibility between the MAE $h_i$ and the feature vector $x_i$ of the $i$-th spatiotemporal segment. In our implementation, rather than using the raw feature~\cite{wang13_iccv}, we use the output of the MAE classifier on the feature vector of spatiotemporal segment $i$. In order to learn biases between different MAEs, we append a constant $1$ to make $x_i$ 2-dimensional. 

{\bf Co-occurrence Model $b_{Y_n,h_i}, b_{h_i,h_j}'$}: This potential captures the co-occurrence constraints between pairs of MAE labels. Since the MAEs are discovered on a per-action basis, thus we restrict the co-occurrence model to allow for only action-consistent types: $b_{Y_n,h_i} = 0$ if the MAE $h_j$ is generated from the action class $Y_n$, and $-\infty$ otherwise. Similarly, $b_{h_i,h_j}' = 0$ if the pair of MAEs $h_i$ and $h_j$ are generated from the same action class, and $-\infty$ otherwise. 

{\bf Spatial-Temporal Model $\beta_{h_i,h_j}^\top d_{ij}$}: This potential captures the spatiotemporal relations between a pair of MAEs $h_i$ and $h_j$. In our experiments, we explore a simplified version of the spatiotemporal model with a reduced set of structures: $\beta_{h_i,h_j}^\top d_{ij} = \beta_{h_i}^\top bin_s(i) + \beta_{h_i}^\top bin_t(i,j)$. The simplification states that the relative spatial and temporal relation of a spatiotemporal segment $i$ with respect to its parent $j$ is dependent on the segment type $h_i$, but not its parent type $h_j$. To compute the spatial feature $bin_s$, we divide a video frame into $5\times 5$ cells, and $bin_s(i)=1$ if the $i$-th spatiotemporal segment falls into the $m$-th cell, otherwise $0$. $bin_t(i,j)$ is a temporal feature that bins the relative temporal location of spatiotemporal segment $i$ and $j$ into one of three canonical relations including \emph{before}, \emph{co-occur}, and \emph{after}. Hence $bin_t(i,j)$ is a sparse vector of all zeros with a single one for the bin occupied by the temporal relation between $i$ and $j$. 

{\bf Root Model $\eta_{Y_n}^\top x_0$}: This potential function captures the compatibility between the global feature $x_0$ of the video $V_n$ and the action class $Y_n$. In our experiment, the global feature $x_0$ is computed as the aggregation of feature descriptors of all spatiotemporal segments in the video. 

\subsection{Inference}
The goal of inference is to predict the hierarchical labeling for a video, including the action label for the whole video as well as the MAE labels for spatiotemporal segments at multiple scales. For a video $V_n$, our inference corresponds to solving the following optimization problem: $(Y_n^*, H_n^*) = \arg\max_{Y_n,H_n} S_{V_n}(X_n, Y_n, H_n)$. For the video $V_n$, we jointly infer the action label $Y_n$ of the video and the MAE labels $H_n$ of the spatiotemporal segments. The inference on the tree structure is exact; and we solve it using belief propagation. We emphasize that our inference returns a parsing of videos including the action label and the MAE labels at multiple levels of granularity. 

\subsection{Learning}
Given a collection of training examples in the form of $\{X_n, H_n, Y_n\}$, we adopt a structured SVM formulation to learn the model parameters $w$. In the following, we develop two learning frameworks for action recognition and parsing respectively. 

{\bf Action Recognition.} We consider a weakly supervised setting. For a training video $V_n$, only the action label $Y_n$ is provided. The MAE labels $H_n$ are automatically discovered using our discriminative clustering algorithm. We formulate it as follows:
\begin{align}
\label{eq:ssvm1}
&\min_{w,\xi\geq 0} \ \frac{1}{2}||w||^2+C\sum_n\xi_n \nonumber\\
&S_{V_n}(X^n,H^n,Y^n)-S_{V_n}(X^n,H^*,Y^*) \nonumber\\
&\geq\Delta_{0/1}~(Y^n,Y^*)-\xi_n, \forall n ,
\end{align}
where the loss function $\Delta_{0/1}~(Y^n,Y^*)$ is a standard $0$-$1$ loss that measures the difference between the ground-truth action label $Y^n$ and the predicted action $Y^*$ for the $n$-th video. We use the bundle optimization solver in~\cite{do09_icml} to solve the learning problem. 

{\bf Action Parsing.} In the real world, a video sequence is usually not bounded for a single action, but may contain multiple actions of different levels of granularity: some actions occur in a sequential order; some actions could be composed of finer-grained MAEs. See Fig.~\ref{fig:vis_action_parsing} for examples. 



The proposed model can naturally be extended for action parsing. Similar to our action recognition framework, the first step of action parsing is to construct the spatiotemporal segment hierarchy for an input video sequence $V_n$, as shown in Fig.~\ref{fig:pipeline}. The only difference is that the input video is not a short video clip, but a long video sequence composed of multiple action and MAE instances. In training, we first associate each automatically discovered spatiotemporal segment with a ground truth action (or MAE) label. If the spatiotemporal segment contains more than one ground truth label, we choose the label with the maximum temporal overlap. We use $Z_n$ to denote the ground truth action and MAE labels associated with the video $V_n$: $Z_n=\{z_i: i=1,\ldots,M_n\}$, where $z_i\in\mathcal{Z}$ is the ground truth action (or MAE) label of the spatiotemporal segment $v_i$, and $M_n$ is the total number of spatiotemporal segments discovered from the video. The goal of training is to learn a model that can parse the input video into a label hierarchy similar to the ground truth annotation $Z_n$. We formulate it as follows:


\begin{align}
\label{eq:ssvm2}
&\min_{w,\xi\geq 0} \ \frac{1}{2}||w||^2+C\sum_n\xi_n \nonumber\\
&S_{V_n}(X^n,Z^n)-S_{V_n}(X^n,Z^*) \geq\Delta~(Z^n,Z^*)-\xi_n, \forall n ,
\end{align}
where $\Delta~(Z^n,Z^*)$ is a loss function for action parsing, which we define as: $\Delta~(Z^n,Z^*)=\frac{1}{M_n}\sum_{i\in\mathcal{V}_n}~\Delta_{0/1}~(z_i^n,z_i^*)$, where $Z^n$ is the ground truth label hierarchy, $Z^*$ is the predicted label hierarchy and $M_n$ is the total number of spatiotemporal segments. Note that the learning framework of action parsing is similar to Eq.~\eqref{eq:ssvm1}, and the only difference lies in the loss function: we penalize incorrect predictions for every node of the spatiotemporal segments hierarchy. 

\section{Experiments}
\label{sec:experiment}
We conduct experiments on both action recognition and parsing. We first describe the datasets and experimental settings. We then present our results and compare with the state-of-the-art results on these datasets. 

\subsection{Experimental Settings and Baselines}
We validate our methods on four challenging benchmark datasets, ranging from fine-grained actions (MPI Cooking), realistic actions in sports (UCF Sports) and movies (Hollywood2) to untrimmed action videos (THUMOS challenge). In the following, we briefly describe the datasets, experimental settings and baselines. 

{\bf MPI Cooking dataset}~\cite{rohrbach12_cvpr} is a large-scale dataset of $65$ fine-grained actions in cooking. It contains in total $44$ video sequences (or equally $5609$ video clips, and $881,755$ frames), continuously recorded in kitchen. The dataset is very challenging in terms of distinguishing between actions of small inter-class variations, e.g. cut slices and cut dice. We split the dataset by taking one third of the videos to form the test set and the rest of the videos are used for training. 


{\bf UCF-Sports dataset}~\cite{rodriguez2008cvpr} consists of $150$ video clips extracted from sports broadcasts. Compared to MPI Cooking, the scale of UCF-Sports is small and the durations of the video clips it contains are short. However, the dataset poses many challenges due to large intra-class variations and camera motion. For evaluation, we apply the same train-test split as recommended by the authors of~\cite{lan11_iccv}. 

{\bf Hollywood2 dataset}~\cite{marszalek09_cvpr} is composed of 1,707 video clips (823 for training and 884 for testing) with 12 classes of human actions. These clips are collected from 69 Hollywood movies, divided into 33 training movies and 36 testing movies. In these clips, actions are performed in realistic settings with camera motion and great variations.

{\bf THUMOS challenge 2014}~\cite{THUMOS14} contains over 254 hours of temporally untrimmed videos and 25 million frames. We follow the settings of the action detection challenge. We use $200$ untrimmed videos for training and $211$ untrimmed videos for testing. These videos contain 20 action classes and are a subset of the entire THUMOS dataset. We consider a weakly supervised setting: in training, each untrimmed video is only labeled with the action class that the video contains, neither spatial nor temporal annotations are provided. Our goal is to evaluate the ability of our model in automatically extracting useful mid-level action elements (MAEs) and structures from large-scale untrimmed data.     


{\bf Baselines.} In order to comprehensively evaluate the performance of our method, we use the following baseline methods. 1) {\em DTF:} the first baseline is the dense trajectory method~\cite{wang13_iccv}, which has produced the state-of-the-art performance in multiple action recognition benchmarks. 2) {\em IDTF:} the second baseline is the improved dense trajectory feature proposed in~\cite{wang13_iccv}, which uses fisher vectors (FV)~\cite{perronnin10_eccv} to encode the dense trajectory features. FV encoding~\cite{wang13_iccv,oneata13_iccv} has been shown an improved performance over traditional Bag-of-Features encoding. 3) {\em root model:} the third baseline is equivalent to our model without the hierarchical structure, which only uses the IDTF features that fall into the spatiotemporal segments discovered by our method, while ignoring those in the background. 4) {\em sliding window:} the fourth baseline runs sliding windows of different lengths and step sizes on an input video sequence, and performs non-maximum suppression to find the correct intervals of an action. This baseline is only applied to action recognition of untrimmed videos and action parsing.

\subsection{Experimental Results}
We summarize the action recognition results on multiple benchmark datasets in Table~\ref{tab:acc_soa},~\ref{tab:acc_mpi} and~\ref{tab:acc_thumos} respectively.

\begin{table}
\small
\centering
\begin{tabular}{|c|c|c|c|} 
\hline
MPI Cooking & Per-Class \\
\hline\hline 
DTF~\cite{rohrbach12_cvpr,wang13_iccv} & 38.5\\
\hline
root model\,(ours) & 43.2\\
full model\,(ours) & {\bf 48.4} \\
\hline
\end{tabular}
\vspace{0.03cm}
\caption{Comparison of action recognition accuracies of different methods on the MPI Cooking dataset.}
\label{tab:acc_mpi}
\end{table}

\begin{table}
\small
\centering
\begin{tabular}{|c|c|c|c|} 
\hline
UCF-Sports & Per-Class & Hollywood2 & mAP \\
\hline\hline 
Lan~et~al.~\cite{lan11_iccv} & 73.1 & Gaidon~et~al.~\cite{gaidon14_ijcv} & 54.4 \\
Tian~et~al.~\cite{tian13_cvpr}  & 75.2 & Oneata~et~al.~\cite{oneata14_cvpr} & 62.4 \\ 
Raptis~et~al.~\cite{raptis12_cvpr} & 79.4 & Jain~et~al.~\cite{jain13_cvpr} & 62.5 \\ 
Ma~et~al.~\cite{ma13_iccv}  & 81.7 & Wang~et~al.~\cite{wang13_iccv} & 64.3 \\
\hline\hline
IDTF~\cite{wang13_iccv} & 79.2 & IDTF~\cite{wang13_iccv} & 63.0 \\ 
\hline
root model~(ours) & 80.8 & root model~(ours) & 64.9 \\ 
full model~(ours) & {\bf 83.6} & full model~(ours) & {\bf 66.3}\\
\hline
\end{tabular}
\vspace{0.1cm}
\caption{Comparison of our results to the state-of-the-art methods on UCF-Sports and Hollywood2 datasets. Among all of the methods,~\cite{raptis12_cvpr},~\cite{ma13_iccv},~\cite{gaidon14_ijcv} and our full model use hierarchical structures.} \label{tab:acc_soa}
\end{table}

\begin{table}
\small
\centering
\begin{tabular}{|c|c|} 
\hline
THUMOS (untrimmed) & mAP \\
\hline\hline 
IDTF~\cite{wang13_iccv} & 63.0 \\ 
\hline
sliding window & 63.8 \\ 
\hline
INRIA (temporally supervised)~\cite{oneata14_thumos} & {\bf 66.3}\\
\hline
\hline
full model~(ours) & 65.4\\
\hline
\end{tabular}
\vspace{0.06cm}
\caption{Comparison of action recognition accuracies of different methods on the THUMOS challenge (untrimmed videos).} \label{tab:acc_thumos}
\end{table}

{\bf Action recognition.} Most existing action recognition benchmarks are composed of video clips that have been trimmed according to the action of interest. On all three benchmarks~(i.e. UCF-Sports, Hollywood2 and MPI), our full model with rich hierarchical structures significantly outperforms our own baseline {\em root model}~(i.e. our model without hierarchical structures), which only considers the dense trajectories extracted from the spatiotemporal segments discovered by our method. We can also observe that the root model consistently improves dense trajectories~\cite{wang13_iccv} on all three datasets. This demonstrates that our automatically discovered MAEs fire on the action-related regions and thus remove the irrelevant background trajectories.

We also compare our method with the most recent results reported in the literature for UCF-Sports and Hollywood2. On UCF-Sports, all presented results follow the same train-test split~\cite{lan11_iccv}. The baseline IDTF~\cite{ma13_iccv} is among the top performance. Ma~et~al.~\cite{ma13_iccv} reported $81.7\%$ by using a bag of hierarchical space-time segments representation. We further improve their results by around $2\%$. On Hollywood2, our method also achieves state-of-the-art performance. The previous best result is from~\cite{wang13_iccv}. We improve it further by $2\%$. Compared to the previous methods, our method is weakly supervised and does not require expensive bounding box annotations in training~(e.g.~\cite{lan11_iccv, tian13_cvpr, raptis12_cvpr}) or human detection as input~(e.g.~\cite{wang13_iccv}). 

\begin{figure}
\centering
\includegraphics[width=0.85\linewidth]{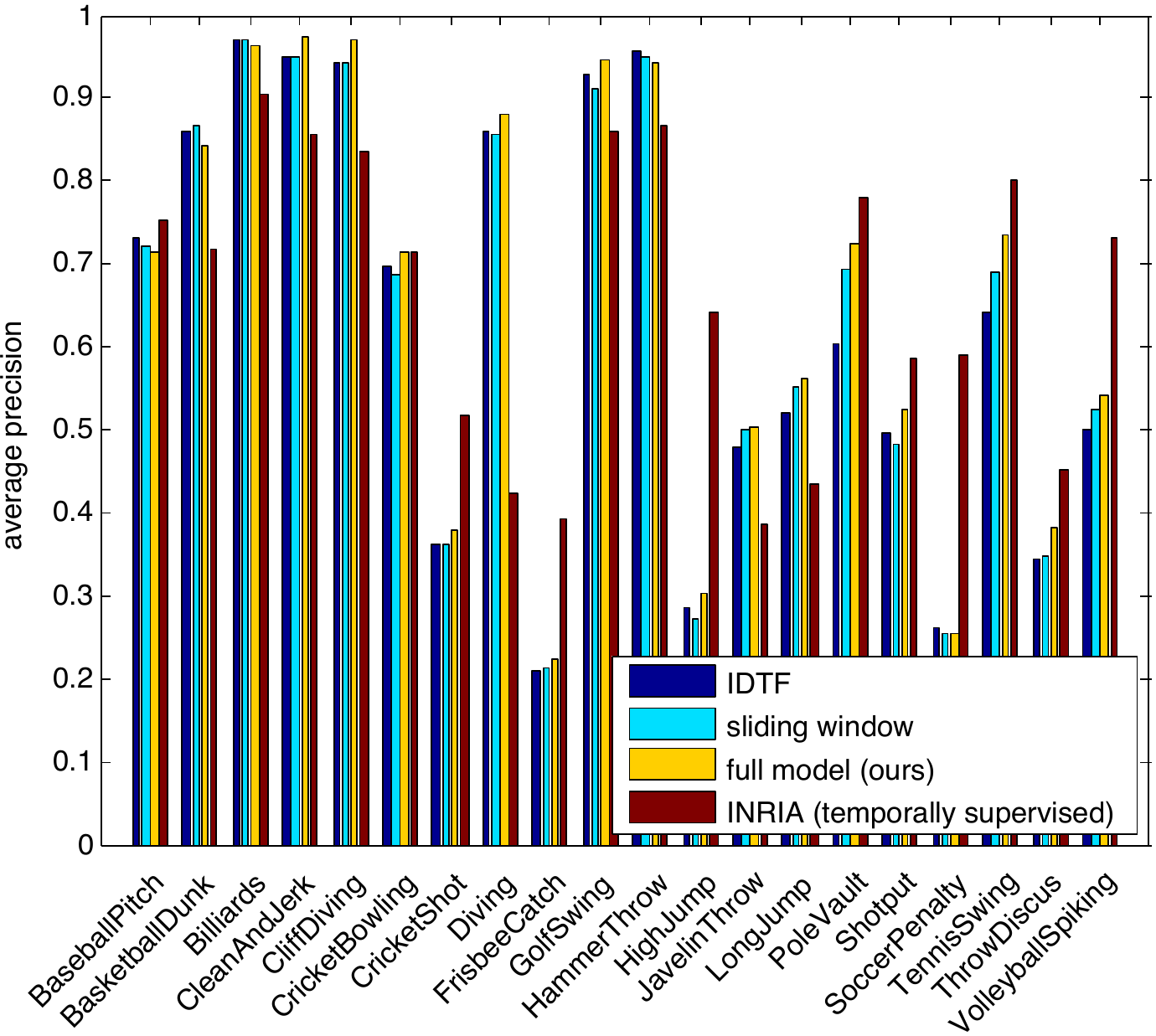}
\caption{Average precisions of the 20 action classes of untrimmed videos from the temporal localization challenge in THUMOS.}
\label{fig:thumos_ap}
\end{figure}

On THUMOS challenge that is composed of realistic untrimmed videos~(Table~\ref{tab:acc_thumos}), our method outperforms both IDTF and the sliding window baseline. Given the scale of the dataset, we skip the time-consuming spatial region proposals and represent action as a hierarchy of temporal segments, i.e. each frame is regarded as a ``spatial segment''. Our method automatically identifies the temporal segments that are both representative and discriminative for each action class without any temporal annotation of actions in training. We also compare our methods with the best submission~(INRIA~\cite{oneata14_thumos}) of the temporal action localization challenge in THUMOS 2014. INRIA~\cite{oneata14_thumos} uses a mixture of IDTF~\cite{wang13_iccv}, SIFT~\cite{lowe99_iccv}, color features~\cite{clinchant2008trans} and the CNN features~\cite{jia2014caffe}. Also, their model~\cite{oneata14_thumos} is \emph{temporally supervised}, which uses temporal annotations (the start and end frames of actions in untrimmed videos) and additional background videos in training. Our method achieves a competitive performance (within 1\%) using only IDTF~\cite{wang13_iccv} and doesn't require any temporal supervision.  We provide the average precisions (AP) of all the 20 action classes in Fig.~\ref{fig:thumos_ap}. Our method outperforms \cite{oneata14_thumos} in 10 out of the 20 classes, especially \emph{Diving} and \emph{CleanAndJerk}, which contain rich structures and significant intra-class variations.

\begin{figure}
\centering
\includegraphics[width=0.8\columnwidth]{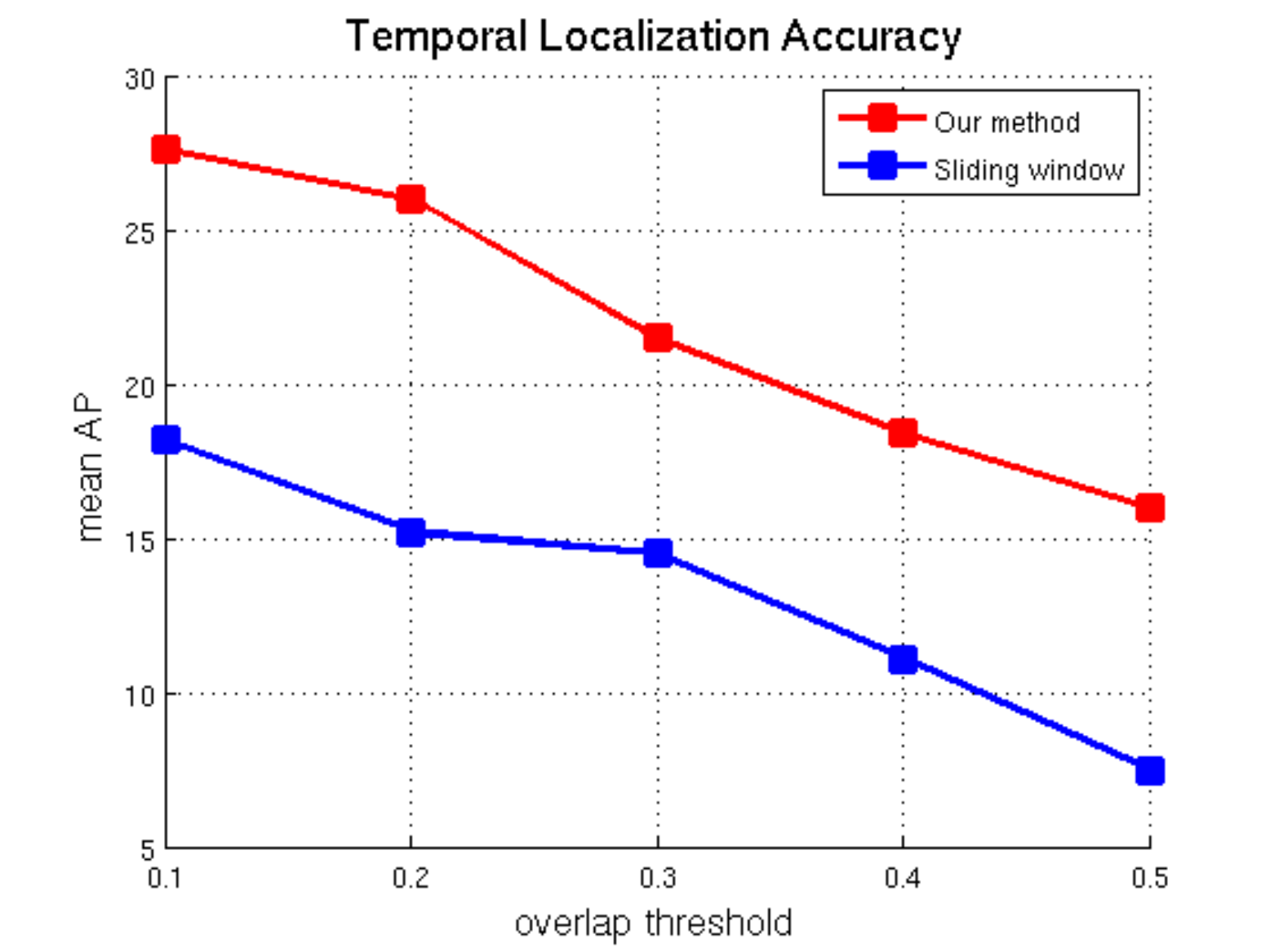} 
\caption{{\bf Action parsing performance.} We report mean Average precision (mAP) of our method and the sliding window baseline on MPI Cooking with respect to different overlapping thresholds that determine whether an action (or MAE) segment is correctly localized.}
\label{fig:parsing_ap}
\end{figure}

\begin{figure*} 
\centering
\includegraphics[width=2\columnwidth]{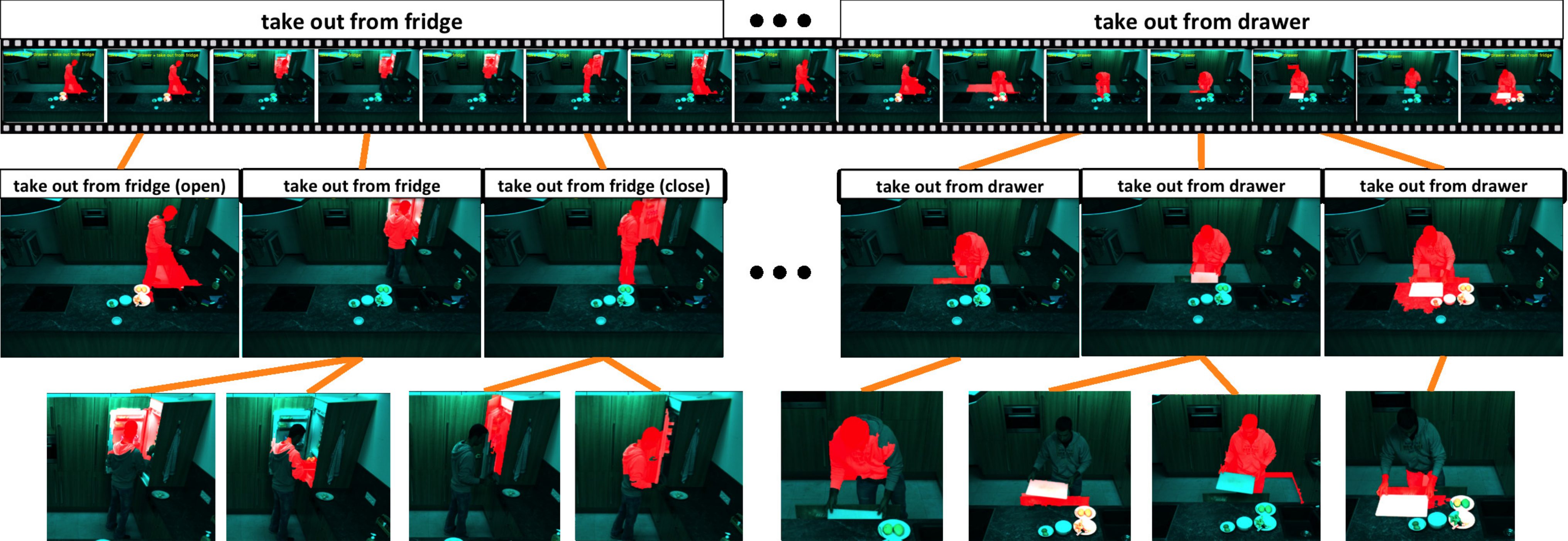} \\
(a) A test video containing ``take out from fridge'' and ``take out from drawer'' actions.\\
\vspace{1mm}
 \includegraphics[width=2\columnwidth]{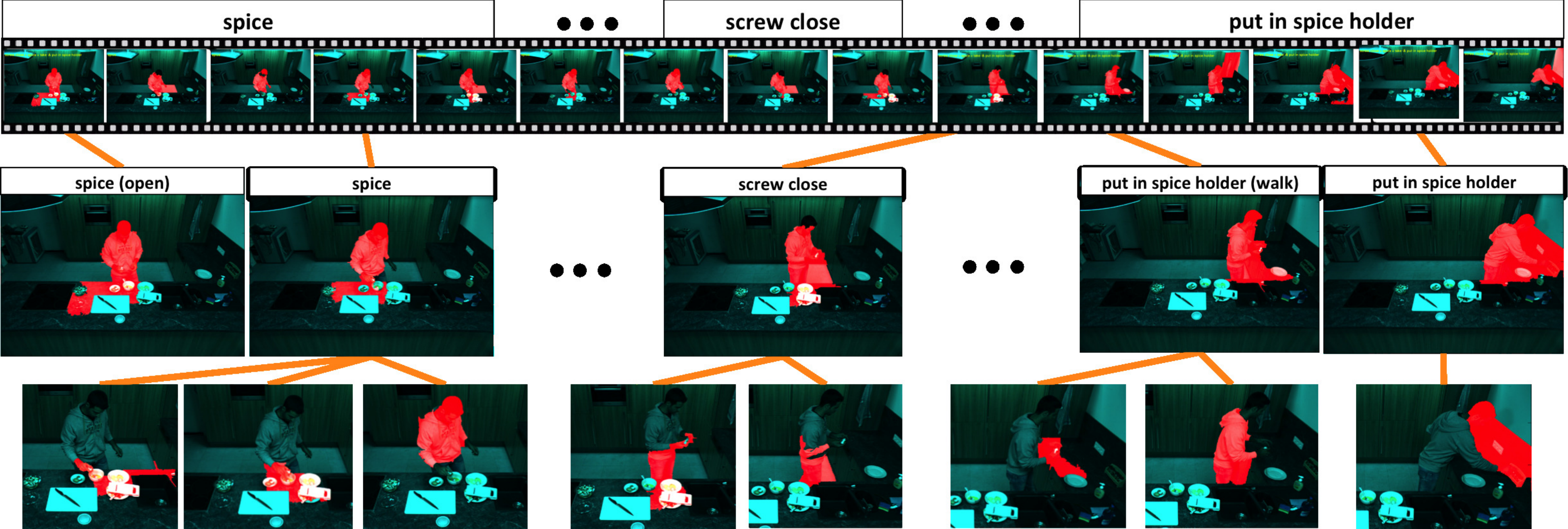} \\
 (b) A test video containing ``spice'', ``screw close'' and ``put in spice holder'' actions.\\
\vspace{1mm}
\caption{{\bf Action parsing.} This figure shows the output of our action parser for two test videos. For each video, we visualize the inferred fine-grained action labels (shown on top of each image), the MAE segments (the red masks in each image) and the parent-child relations (the orange line). As we can see, our action parser is able to parse long video sequences into representative action patterns (i.e. MAEs) at multiple scales. Note that the figure only includes a few representative nodes of the entire tree obtained by our parser, we provide more visualizations in the supplementary material. 
}
\label{fig:vis_action_parsing}
\end{figure*}

{\bf Action parsing.} Given a video sequence that contains multiple action and MAE instances, our goal is to localize each one of them. Thus during training, we assume that all of the action and MAE labels as well as their temporal extent are provided. This is different from action recognition where all of the MAEs are unsupervised. We evaluate the ability of our method to perform action parsing by measuring the accuracy in temporally localizing all of the action and MAE instances. An action (or MAE) segment 
is considered as true positive if it overlaps with the ground truth segment beyond a pre-defined threshold. We evaluate the mean Average Precision (mAP) with the overlap threshold varying from $0.1$ to $0.5$.

We use the original fine-grained action labels provided in the MPI Cooking dataset as the MAEs at the bottom level of the hierarchy, and automatically generate a set of higher-level labels by composing the fine-grained action labels. The detailed setups are explained in the supplementary material. Examples of higher-level action labels are: ``cut apart - put in bowl'', ``screw open - spice - screw close''. We only consider labels with length ranging from $1$ to $4$, and occur in the training set for more than $10$ times. In this way, we have in total $120$ action and MAE labels for parsing evaluation.



We compare our result with the sliding window baseline. The curves are shown in Fig.~\ref{fig:parsing_ap}. Our method shows consistent improvement over the baseline using different overlap threshold. If we consider an action segment is correctly localized based on ``intersection-over-union'' score larger than $0.5$~(the PASCAL VOC criterion), our method outperforms the baseline by $8.5\%$. The mean performance gap (averaged over all different overlap threshold) between our method and the baseline is $8.6\%$. Some visualizations of action parsing results are shown in Fig.~\ref{fig:vis_action_parsing}. As we can see, the story of human actions is more than just the actor: as shown in the figure, the automatically discovered MAEs cover different aspects of an action, ranging from human body and parts to spatiotemporal segments that are not directly related to humans but carry significant discriminative power (e.g. a piece of fridge segment for the action ``take out from fridge''). This diverse set of mid-level visual patterns are then organized in a hierarchical way to explain the complex store of the video at different levels of granularity.  

\section{Conclusion}
\label{sec:conclusion}
We have presented a {\em hierarchical mid-level action element (MAE)} representation for action recognition and parsing in videos. We consider a weakly supervised setting, where only the action labels are provided in training. Our method automatically parses an input video into a hierarchy of MAEs at multiple scales, where each MAE defines an action-related spatiotemporal segment in the video. We develop structured models to capture the rich semantic meanings carried by these MAEs, as well as the spatial, temporal and hierarchical relations among them. In this way, the action and MAE labels at different levels of granularity are jointly inferred. Our experimental results demonstrate encouraging performance over a number of standard baseline approaches as well as other reported results on several benchmark datasets. 

{\small
\bibliographystyle{ieee}
\bibliography{lan}
}

\end{document}